\documentclass[10pt,twocolumn,letterpaper]{article}

\usepackage{cvpr}
\usepackage{times}
\usepackage{epsfig}
\usepackage{graphicx}
\usepackage{amsmath}
\usepackage{amssymb}
\usepackage[font=small,skip=-2pt]{caption}
\cvprfinalcopy 

\def\cvprPaperID{5560} 

\ifcvprfinal\fi
\begin{document}

\title{An End-to-End Network for Generating Social Relationship Graphs}

\author{\parbox{16cm}{\centering
    {\large Arushi Goel$^1$, Keng Teck Ma$^{1,2}$, and Cheston Tan$^{2}$}\\
    {\normalsize
    $^1$A*STAR Artificial Intelligence Initiative, Singapore, 
    $^2$Institute for Infocomm Research, A*STAR, Singapore}} \\
    {{\tt\small goela@ihpc.a-star.edu.sg}, {{\tt\small ma\_ken\_teck@scei.a-star.edu.sg}}, {{\tt\small cheston-tan@i2r.a-star.edu.sg}}}
}

\maketitle

\begin{abstract}
   Socially-intelligent agents are of growing interest in artificial intelligence. To this end, we need systems that can understand social relationships in diverse social contexts. Inferring the social context in a given visual scene not only involves recognizing objects, but also demands a more in-depth understanding of the relationships and attributes of the people involved. To achieve this, one computational approach for representing human relationships and attributes is to use an explicit knowledge graph, which allows for high-level reasoning. We introduce a novel end-to-end-trainable neural network that is capable of generating a Social Relationship Graph -- a structured, unified representation of social relationships and attributes -- from a given input image. Our Social Relationship Graph Generation Network (SRG-GN) is the first to use memory cells like Gated Recurrent Units (GRUs) to iteratively update the social relationship states in a graph using scene and attribute context. The neural network exploits the recurrent connections among the GRUs to implement message passing between nodes and edges in the graph, and results in significant improvement over previous methods for social relationship recognition.
\end{abstract}

\begin{figure}[!tb]
\begin{center}
\includegraphics[width=0.8\linewidth]{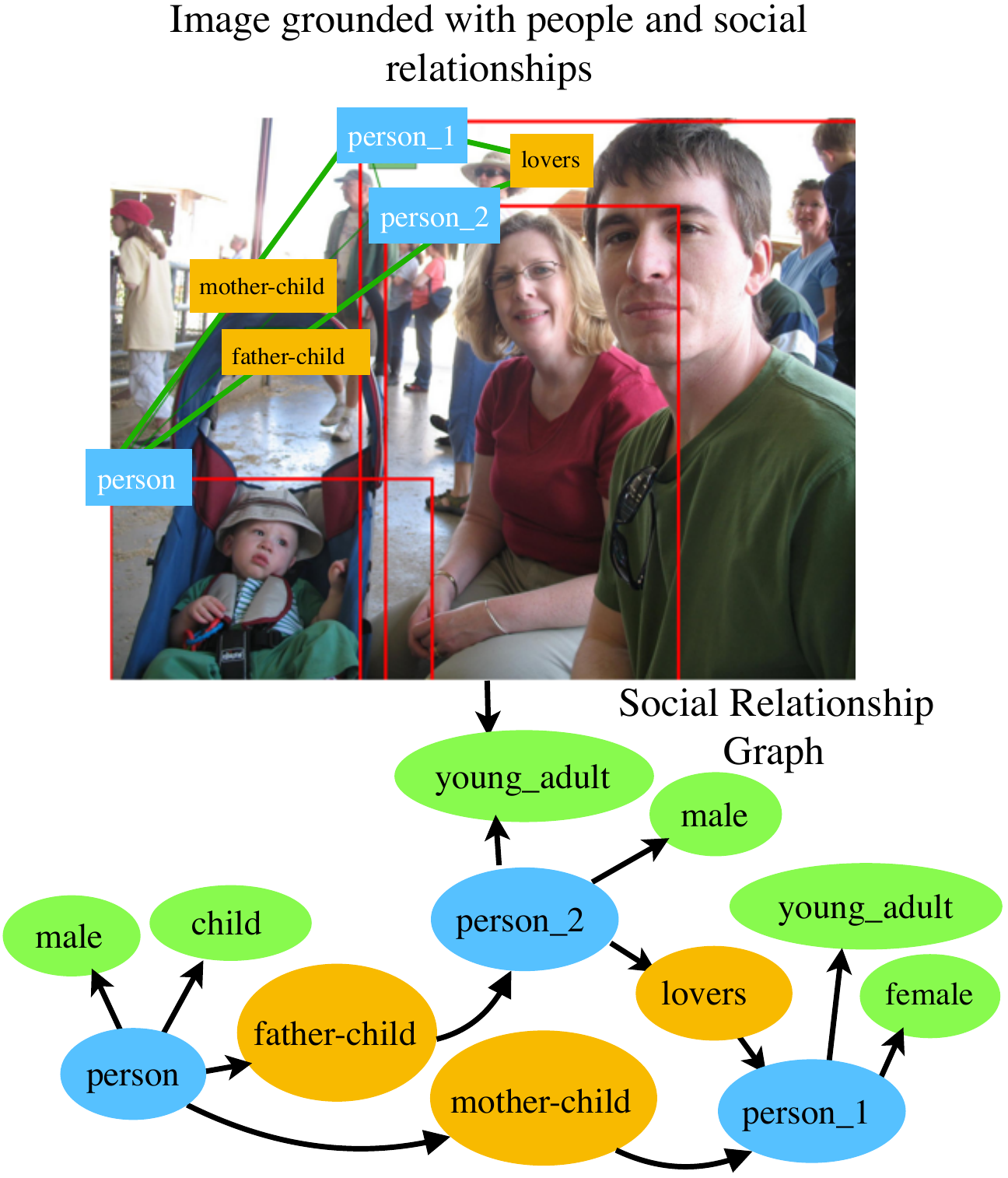}
\end{center}
\caption{For a given scene, our network generates a structured representation -- a Social Relationship Graph. Graph representations have shown good results on a variety of high-level vision tasks, e.g. image retrieval and visual Q\&A. }
\label{fig:SRG}
\end{figure}

\section{Introduction}

The understanding of human relationships in computer vision research is in its nascent stage. In comparison, significant efforts have been made by social psychologists and other researchers to study social relationships in humans \cite{frith2009role,li2017dual}. The pioneering work of Sun \etal \cite{SRR} proposes a social relationship framework based on Bugental's Social Domain Theory \cite{Bugental} to classify social relationships and domains. In this paper, we take a step further in understanding social relationships from images by generating a \textit{Social Relationship Graph (SRG)}, as illustrated in Figure \ref{fig:SRG}.

In recent computer vision research, predicting relationships of the ``subject-predicate-object'' kind have gained major research attention. These can be used for multiple high-level tasks like image retrieval, image captioning, and visual question answering \cite{Retrieval,QA,Caption}. The recent work for the generation of scene graphs using an end-to-end model \cite{MP,SGG,zellers2018neural} gives the best results on the Visual Genome Dataset \cite{VG}. Since such graphs are human-interpretable, we propose to build a Social Relationship Graph, which encodes relationship and attribute information and captures the rich semantic structure of a scene.

The task of understanding human relationships is a challenging problem given the wide variations that humans pose in their environments. There is unobservable, latent information in images which we as humans find easy to interpret. For developing human-level understanding in such situations, computational models are based on the theories of social and cognitive psychology \cite{smith2000dual}. Based on the social psychology theories of Bugental \cite{Bugental}, we focus on human attributes and environments for social relationships.

Scene and global contextual cues have the best results for social relationships \cite{li2017dual}. Furthermore, the activity that people are partaking in provides crucial features for social relationship classification \cite{SRR}. In social psychology research \cite{Bugental}, it has been shown that appearance cues such as age, gender and clothing are useful in understanding social relationships. We thus use scene context, activity and appearance features for social relationship graph inference.

We formulate our problem as graph inference that encodes the interactions between nodes and edges in a graph. Our problem is more challenging than scene graph generation \cite{MP,SGG,zellers2018neural} as our work requires understanding of high-level social semantic features (\textit{e.g. social context}) and
low-level visual features (\textit{e.g. spatial arrangement of objects}). 
 
We devise a novel end-to-end model for predicting social relationships using a \textit{Social Relationship Graph Generation Network (SRG-GN)} that combines inputs from a \textit{Multi-Network Convolutional Neural Network (MN-CNN)} to iteratively update the hidden states of the nodes (persons) and edges (relationships) in a \textit{Social Relationship Graph Inference Network (SRG-IN)} by passing messages between two types of Gated Recurrent Units (GRUs) \cite{chung2014empirical}.

The \textit{Rship GRUs} (edges) have the scene and activity features as the input, while the \textit{PPair GRUs} (nodes) have the human attribute features as input. The hidden state for each edge gets updated by combining the updated node state and updated edge state. Thus, the relationship (edge) state gets updated by the fine-grained attribute features of the adjacent nodes and the scene and activity context from nearby edges. 

The main contributions of this paper are: 1) a novel structured representation (Social Relationship Graph) for social understanding in visual scenes; 2) a novel end-to-end-trainable neural network architecture using GRUs and semantic attributes for graph-generation; 3) new state-of-the art results for social relationship recognition on the PIPA-relation \cite{SRR} and PISC \cite{li2017dual} datasets. This is the first architecture that builds on social relationships and attributes using memory cells, and our results demonstrate the importance of message passing and scene context.
\section{Related Work}
\subsection{Social Relationship Recognition}

The area of social relationships is of growing interest to the community, as social chatbots and personal assistants need to understand social interactions. Many researchers have tried to understand social relationships, roles and interactions. Zhang \etal \cite{zhang2015learning} have studied interpersonal relationships using facial expressions with a Siamese-like architecture. There are studies on Kinship recognition \cite{Kinrec} and Kinship verification \cite{fang2010towards}. Wang \etal \cite{wang2010seeing} studies family relationships in personal image collections. Jinna \etal \cite{multistream} introduced a video dataset for coarse-grained social relationships between humans. Li \etal \cite{li2017dual} predicts social relationships in images using an Attentive-RCNN model for 6-relationship categorization. Ramanathan \etal \cite{ramanathan2013social} recognize social roles played by people in various events. Chakraborty \etal \cite{chakraborty20133d} classify photos into classes such as `couple, family, group, or crowd'. Sun \etal \cite{SRR} predict social relationships for fine-grained relationships between humans in everyday images. Many of the above-mentioned works have used physical appearance or cues like activity, proximity, emotion, expression, context etc. Our work differs by combining the essential attribute features with memory cells providing a richer framework for our problem. 

\subsection{Graph-Based Representations}
There is a lot of recent interest in using structured graph representations for visual grounding of images. Knowledge graphs are being widely used for object detection and image classification \cite{fang2017object,marino2016more}. Johnson \etal \cite{Retrieval} introduced ground-truth annotated scene graphs for the task of image retrieval using object relationships and attributes. 
Since then, the task of generating scene graphs directly from images by using intrinsic graph properties and surrounding context has gained attention \cite{MP,SGG,zellers2018neural,herzig2018mapping}. 
The use of vision and language modules together has also been explored by researchers for identifying relationships between objects \cite{lu2016visual}. 
We present a novel framework for generating graphs, focusing on social relationships and attributes of people, unlike the focus on spatial object relationships in existing work.
\begin{figure*}[!tb]
\begin{center}
\includegraphics[width=\linewidth]{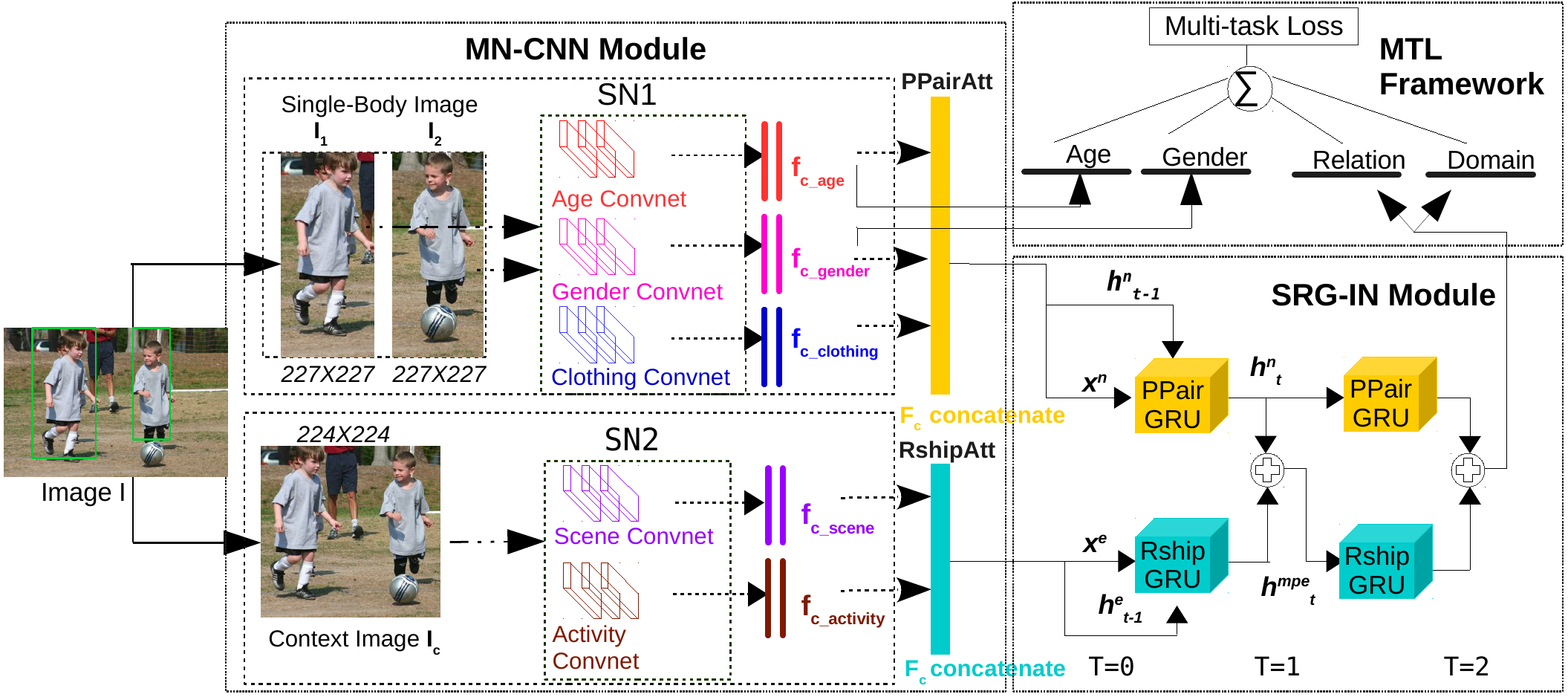}
\end{center}
   \caption{SRG-GN: Our proposed end-to-end network for Social Relationship Graph generation. We take the single body images, \textit{$I_1$} and \textit{$I_2$}, and the ``context image'' (smallest image that contains both single-body images), $I_c$ as input to the SN1 and SN2 sub-modules of the MN-CNN module and fine-tune the fully-connected layers of all the attributes. These fully-connected layers are concatenated and fed as input to the SRG-IN module and the hidden edge state gets iteratively updated by mean-pooling the edge (relation) and node (person/attribute) hidden states. The final updated edge state is used for predicting social relationships in the given image. For the multi-task learning framework, age and gender attributes from the fully-connected layers of the MN-CNN module also contribute to the joint optimization of the individual cross-entropy losses. The symbol $\sum$ denotes summation and $\bigoplus$ denotes mean-pooling.}
\label{fig:endtoend}
\end{figure*}

\section{Model Definition}
\label{model}
In this section, we provide an overview of our method for generating Social Relationship Graphs from images using our Social Relationship Graph Generation Network (SRG-GN). The framework in Figure \ref{fig:endtoend} gives a more detailed description of our two modules: A Multi-Network Convolutional Neural Network (MN-CNN) module for Attribute and Relationship representations followed by a Social Relationship Graph Inference Network (SRG-IN) module for generating a structured graph representation. The model is trained end-to-end to predict relationships, domains and attributes as part of a scene in the form of a structured semantic directed graph representation. 

\subsection{Multi-Network Convolutional Neural Network (MN-CNN) for Relationships and Attributes}

We have an input image \textbf{I} and a set of bounding box annotations ${B_i}$ for the people in image \textbf{I} where \textit{i = 1,2,...,N}. These annotations are cropped for a single-body image of a person, ${I_i}$ and resized into \textit{227x227} pixels. For every annotated relationship between two people, we define a ``context image'' (smallest image that contains both single-body images) \textbf{${I_c}$}, resized into \textit{224x224} pixels.

The MN-CNN module has two sub-modules (SN1 and SN2) with the inputs ${I_i}$ and ${I_c}$ respectively. ${I_i}$ is passed through the sub-module, SN1, which is an Attribute ConvNet architecture with 5 conv layers and 2 fully-connected layers (fc6 and fc7), each for the 3 attributes -- age, gender and clothing. The weights for these 3 ConvNet layers are the pre-trained weights as discussed later in Section \ref{implementation}. We fine-tune the fully-connected layers for each attribute and then the features from the fc7 layers are concatenated into a single feature vector, which we assign to \textit{PPairAtt}.
\begin{equation}
     PPairAtt = [{f_c}_{age|4096d},{f_c}_{gender|4096d},{f_c}_{clothing|4096d}]
\end{equation}
The sub-module SN2 is a network of pairwise-relationship ConvNet architectures. There are two VGG-16 architectures \cite{simonyan2014very} to compute activity and scene features from the context images of people. Activity has an important correlation to identifying relationships between people, say, two people ``marrying'' are more likely to be lovers. Scene context information can also be leveraged for improving the model efficacy to predict relationships. 
As humans too, we understand images by looking at the whole image scene and not only the objects under consideration. This gives more coarse-grained information to comprehend the given task. 
We fine-tune the fully-connected layers for both of these sub-architectures, then concatenate the fc7 layers to form a high-dimensional vector, which we assign to \textit{RshipAtt}.
\begin{equation}
    RshipAtt = [{f_c}_{activity|1024d},{f_c}_{scene|4096d}]
\end{equation}

\subsection{Social Relationship Graph Inference Network (SRG-IN)}

We formulate the task of classifying social relationships between people in the form of a social graph inference problem, where we predict the relationships in an image by considering relationship triplets \textit{$<$person1, relation, person2$>$}. 
Consider a pair of people in the given image \textbf{I} with some social relationship between them. In our network, each relationship in an image gets information from its nearby nodes (person attributes) and also its nearby edges (relationships). 
This is achieved by using Gated Recurrent Units (GRUs) to aggregate messages from the adjacent nodes and relationships and iteratively update those messages to improve the predicted edge states (relationships) between the given nodes (persons). 
Thus, we are able to exploit the information in the scene context and the individual attributes to improve the relationships in the Social Relationship Graph. 

\vspace{-0.4cm}
\subsubsection{Inference using GRUs and Message Passing Scheme:}
Mathematically, we formulate our inference task as a probability function: given an input image \textbf{I}, bounding box values ${B_i}$ and x as the representation of the SRG: 
\begin{equation}
    x = {\{x^{age}_i,x^{gender}_i,x^{relation}_{i->j},I| i = 1,2,...N,j=1,2,...N\}}
\end{equation}

where $x^{age}_i$ and $x^{gender}_i$ are the age and gender attributes of the person and $x^{relation}_{i->j}$ is the social relationship between the persons i and j, and N is the total number of people in an image. We have to find an optimal value of x, 
\begin{equation}
    x^* = argmax_xPr(x|I,B_i)
\end{equation}
where, 
\begin{equation}
    Pr(x|I,B_i) = \prod_{i=1}^{N} \prod_{j=1}^{N} Pr(x^{age}_i,x^{gender}_i,x^{relation}_{i->j}|I,B_i)
\end{equation}
We perform this inference using an end-to-end network of Social Relationship Graph Generation where the \textit{MN-CNN module} provides the initial inputs for the nodes and the edges in the \textit{SRG-IN module}. 

Gated Recurrent Units (GRUs) are the most reliable and lightweight RNN memory units. The GRUs operate using a reset gate and an update gate and have the ability to keep memory from previous activations allowing them to remember features for a long time. Let us briefly revisit the functioning of a single GRU cell. The reset gate \textit{r} is defined as 
\begin{equation}
    r_t = \sigma(W_r.[h_{t-1},x_t])
\end{equation}
where $\sigma$ is the sigmoid function, $W_r$ is the learnable weight matrix, $h_{t-1}$ is the previous hidden state, $x_t$ is the input to the GRU cell and [,] denotes concatenation. The update gate \textit{z} is given by
\begin{equation}
    z_t = \sigma(W_z.[h_{t-1},x_t])
\end{equation}
The actual activation in the memory unit is given by
\begin{equation}
    h_t = (1-z_t)*h_{t-1} + z_t*\tilde{h_t}
\end{equation}
where, 
\begin{equation}
    \tilde{h_t} = tanh(Wx_t + U(r_t*h_{t-1}))
\end{equation}

W and U are weight matrices that are learned and * is the element-wise multiplication. As empirically evaluated \cite{chung2014empirical}, the reset gate \textit{r} sits between the previous activation and the next candidate activation to forget the previous state, and the update gate \textit{z} decides how much of the candidate activation to use in updating the cell state.

Our network has two sets of GRUs (Relationship(\textit{Rship}) and Person-Pair(\textit{PPair})). The initial state of the GRUs can be set to zero or some random vector, and the input to the unit is a sequence of features or symbols. To compute activations from the PPair GRU, we take the feature vector, \textit{PPairAtt}, from the SN1 sub-module of the MN-CNN module as the initial state and input to the PPair GRU. We concatenate the features from the two nodes (persons) with a relationship and take this integrated message as input. To compute activations from the Rship GRU, we take the feature vector, \textit{RshipAtt}, from the SN2 sub-module of the MN-CNN as the initial state and input to the Rship GRU. When the state of the PPair GRU is updated, we update the state of the Rship GRU by including the node state information into the edge state information to provide context to the edges from its adjacent nodes.

Each of the two GRUs receives incoming messages and we concatenate these messages using a standard pooling operation, mean pooling. Mean pooling aggregates messages in a more meaningful representation as shown in Section \ref{pooling}. The PPair GRU receives \textit{[$f_i$,$f_j$]} as input, \textit{$x_n$} where, $f_i$ and $f_j$ are the attribute features of the nodes i and j respectively and [,] denotes concatenation. The previous node state \textit{$h^n_{t-1}$} is also initialized using \textit{[$f_i$,$f_j$]} and updates the node state to \textit{$h^n_t$} using \textit{$x_n$} as input. The Rship GRU receives \textit{$f_{i->j}$} as input, \textit{$x_e$} where, $f_{i->j}$ are the relationship features from the MN-CNN module.  The previous edge state \textit{$h^e_{t-1}$} is initialized using \textit{$f_{i->j}$} and the edge state is updated to the "mean-pooled" edge state, \textit{$h^{mpe}_{t}$}, given by:
\begin{equation}
    h^{mpe}_{t} = \frac{h^e_{t}+h^n_{t}}{2}
\end{equation}

This includes the semantic node information into the edge context for updating the edge state with meaningful information from the adjacent nodes and edges. In the next iteration of the GRU, the input to the GRUs are messages from the previous time step. The updated edge representations are used to predict the relationships between nodes. 

\subsection{Multi-Task Learning (MTL) Framework}

In Multi-Task Learning, we simultaneously learn multiple tasks with some shared layers except for one task-specific layer. This can be achieved if the same dataset has multiple labels for learning. For our problem, we have four task labels (age, gender, domain and relationship) that can be learned using the same network. We jointly optimize the loss function by combining the individual loss functions for all these four tasks. We learn the domain labels together with the relationship labels, so that the network can share some relevant information between these two tasks to improve the overall loss function. For instance, the ``Reciprocity Domain'' refers to relationships that have a reciprocal nature, such as, ``friends'', ``siblings'' and ``classmates''. The output from the Rship GRUs are used to predict the domain and relationship labels, whereas the \textit{${f_c}_{age}$} and the \textit{${f_c}_{gender}$} feature vectors from the MN-CNN module are used to predict the age and gender attribute labels respectively using a cross-entropy loss function. We only consider age and gender attribute predictions because the dataset is limited to only these two attributes. Figure \ref{fig:endtoend} shows how we incorporate the MTL framework in our SRG-GN model.

\section{Empirical Evaluation}
In this section, we evaluate the performance of our model using qualitative and quantitative analysis.
\subsection{Dataset Preparation}

The PIPA-relation dataset \cite{SRR} has 16 fine-grained relationship categories \footnote{\textit{father-child, mother-child, grandpa-grandchild, grandma-grandchild, friends, siblings, classmates, lovers/spouses, presenter-audience, teacher-student, trainer-trainee, leader-subordinate, band members, dance team members, sport team members and colleagues}}. We extend their dataset to a PIPA-relation graph dataset. We expand the ground-truth annotations for faces in PIPA into full human body annotations by following the body proportion measurements; 3 x face width and 6 x face height. This gives us ground-truth annotations for single-body images. The context images are cropped from the full images using bounding box values of the people with relationship annotations. We construct our PIPA-relation graph dataset using two attributes (age and gender) from the attribute annotations published on the PIPA dataset \cite{PersonRI}. The train/val/test set has 6289 images with 13,672 relationships and 16,145 attributes, 270 images with 706 relationships and 753 attributes, 2649 images with 5075 relationships and 6655 attributes.

We further validate the performance of our model on the large--scale People in Social Context (PISC) dataset released by Li \etal \cite{li2017dual}. The PISC dataset has 22,670 images where the person pairs are annotated for 3 coarse-grained relationships (\textit{intimate, not-intimate and no relation}) and 6 fine-grained relationships (\textit{commercial, couple, family, friends, professional and no-relation}). The train/val/test set consist of 16,828 images with 55,400 relationship instances, 500 images and 1,505 instances, 1,250 images and 3,961 instances, respectively.

\subsection{Baselines}
\textbf{Comparison models for PIPA-relation dataset: }
Our baselines are the two end-to-end models trained on the PIPA-relation dataset by Sun \etal \cite{SRR} and the end-to-end model for Scene Graph Generation by Xu \etal \cite{MP} as below:

\textit{Double-Stream (DS) CaffeNet:} Trained from scratch on the entire dataset using a two stream network for each single body of a person to predict relationships between them. 

\textit{Finetuned model from pre-trained on Imagenet:} Uses fixed weights of the conv layers from the Imagenet pre-trained weights and fine-tuned the fully-connected layers on the PIPA-relation dataset. 

\textit{Primal-Dual graph model:} Trained the primal-dual graph model \cite{MP} on the PIPA-relation graph dataset.

\textbf{Comparison models for PISC dataset: }
We compare our models with the models proposed by Li \etal \cite{li2017dual}. An overview of the baseline models by \cite{li2017dual} is given below:

\textit{Pair--CNN+BBox:} Two CNNs for each cropped person image with geometry bounding box features.

\textit{Pair--CNN+BBox+Union:} Pair--CNN+BBox with a single CNN for union region of interest features.
 
\textit{Pair--CNN+BBox+Global:} Pair--CNN+BBox with the whole image as context.

\textit{Pair--CNN+BBox+Scene:} Pair--CNN+BBox with scene features as context.


\textit{Dual-Glance:} Combines Pair--CNN+BBox+Union with attention from contextual information to refine predictions.

\subsection{Implementation Details}
\label{implementation}
The pre-trained weights for age, gender, clothing and activity models are publicly available \cite{SRR}. The pre-trained weights for the Scene ConvNet architecture are from the models published by Zhou \etal \cite{zhou2014learning}. We freeze the weights for all the layers and only fine-tune the fully-connected layers of the MN-CNN module, and the GRUs. The output of both the GRUs have a dimension of 512. A softmax layer computes the final scores for age and gender attributes, domains and relationship labels. In case of PISC dataset, we only get scores for domain and relationships as there are no labels for attributes. We sum all the losses and jointly optimize the total weighted loss, as part of the MTL framework. A learning rate of $10^{-6}$ and 2 time-steps for the GRU are used to train the model. To prevent over-fitting, methods like early-stopping, dropout and regularization are employed. Our model is implemented using Tensorflow \cite{abadi2016tensorflow}.

\begin{table}[!tb]

\begin{center}
\centering
\resizebox{0.8\columnwidth}{!}{
    \begin{tabular}{|c|c|}
    \hline
    MODEL & Accuracy\\
    \hline\hline
    Double-Stream Caffenet & 34.40\%  \\
    Primal-Dual model (Our trained) & 44.91\% \\
    Fine-tuned pre-trained on Imagenet & 46.20\% \\ \hline
    Our MN-CNN module only & 49.75\%  \\
    Our SRG-GN without Scene & 51.79\% \\
    Our SRG-GN (final model) & \textbf{53.56\%} \\
    \hline
    \end{tabular}
    }
\end{center}
\caption{Accuracy for the task of Social Relationship Recognition (\textit{SRRec} on PIPA-relation graph dataset). Chance-level accuracy is 6.25\% (1 in 16).}
\label{table:table2}
\end{table}

\begin{figure*}[t]
\begin{center}
\includegraphics[width=0.95\linewidth]{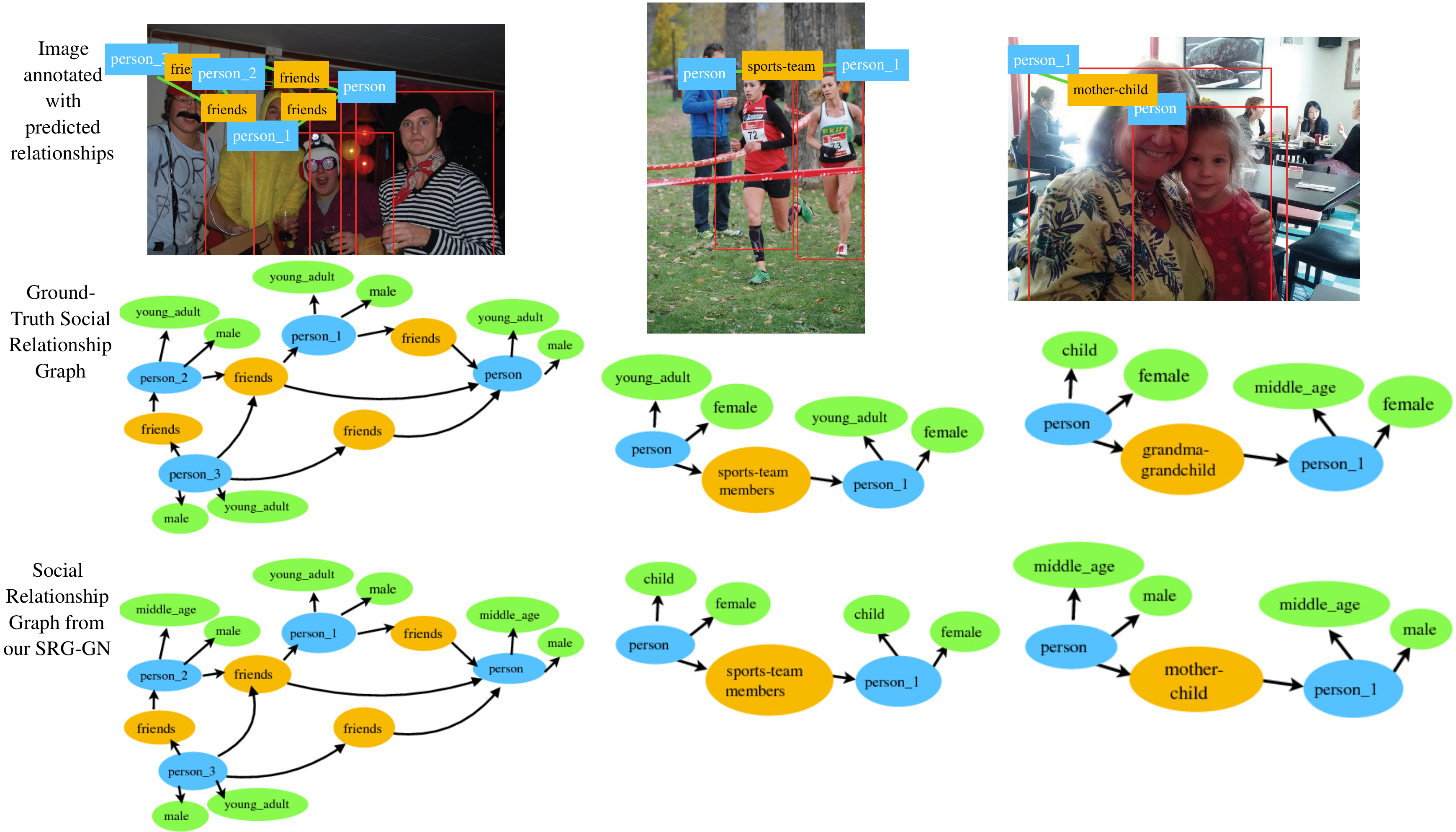}
\end{center}
   \caption{Example Social Relationship Graph generation results from our final model on PIPA-relation graph dataset, and comparison with ground-truth social relationship graphs. Each person (blue ovals) has related age and gender attributes (green ovals) with social relationships between each pair of persons (orange ovals).}
\label{fig:SRG_egs}
\end{figure*}

\subsection{Results}
\label{results}
We evaluate the performance of our model on the PIPA-relation graph dataset and the PISC dataset. The PIPA-relation graph dataset additionally has 6 age labels (\textit{infant, child, young adult, middle age, senior and unknown}) and 2 gender labels (\textit{male and female}).

\subsubsection{Quantitative Results}
We evaluate our model for two setups: 

\textbf{Social Relationship Recognition (SRRec):} To evaluate this, we only consider the triplet predictions of \textit{person-relationship-person} and calculate the accuracy score for social relationship recognition.

\textbf{Social Relationship Graph Generation (SRGGen):} We consider two triplet predictions (\textit{person-relationship-person}; \textit{person-age-gender}) to measure the accuracy of generating a full SRG with correct age and gender nodes and relationship edges.

We report results for different variations of our model and compare with the baselines. \textit{Our MN-CNN module only}, is a variation of our model without the GRUs by using concatenated \textit{PPairAtt} and \textit{RshipAtt} as input to the relationship and domain prediction task specific layers and ${f_c}_{age}$ and ${f_c}_{gender}$ to the age and gender prediction task layers respectively. \textit{Our SRG-GN without scene}, is our final model without the scene context features ${f_c}_{scene}$, in the \textit{RshipAtt}. \textit{Our SRG-GN}, is the final model as shown in Figure \ref{fig:endtoend}.   

\textbf{Results on PIPA-relation dataset:} In Table \ref{table:table2}, we provide the accuracy for our first setup, \textit{SRRec}.
Our \textit{MN-CNN module} improves on the Fine-tuned model by 3.5\% for the task of social relationship recognition. This clearly indicates the importance of using the semantic attributes, scene and activity features over the visual features pre-trained on Imagenet. Our final model, \textit{SRG-GN}, outperforms \textit{only MN-CNN} by 3.81\%, which explains the capability of our message passing scheme for generating social relationship graphs. This technique helps to retain significant information from the nearby nodes and edges in a social relationship graph and thus gives better results. SRG-GN performs better than the primal-dual graph baseline as the latter localizes objects using visual cues with an exchange of information between multiple classes of objects unlike our problem. 
\begin{table*}[!tb]
\begin{center}
\centering
\resizebox{0.8\textwidth}{!}{
\begin{tabular}{|c|c|c|c|c|c|c|c|}
\hline
MODEL & mAP &Family & Couple & Commercial & No-Relation & Professional & Friends  \\
\hline\hline
Our MN-CNN module only & 60.2 & 75.0 & 57.1 & 62.5 & 59.9 & 80.6 & \textbf{26.0} \\
Our SRG-GN without Scene & 69.2 & 80.0 & 77.7 & \textbf{88.8} & 61.7 & \textbf{81.8} & 24.5 \\
Our SRG-GN (final model) & \textbf{71.6} & \textbf{80.0} & \textbf{100.0} & 83.3 & \textbf{62.5} & 78.4 & 25.2 \\
\hline
\end{tabular}}
\end{center}
\caption{Detection results for 6-relationship labels on PISC dataset.}
\label{table:pisc_ap}
\end{table*}
\begin{table}[!tb]

\begin{center}
\centering
\resizebox{0.7\columnwidth}{!}{
\begin{tabular}{|c|c|c|}
\hline
MODEL & Accuracy \\
\hline\hline
Our SRG-GN without Scene &  20.24\% \\
Our SRG-GN (final model) & \textbf{27.64\%} \\
\hline
\end{tabular}
}
\end{center}
\caption{Accuracy for the task of Social Relationship Graph Generation (\textit{SRGGen}) on PIPA-relation graph dataset. Chance- level accuracy is 0.52\% = (1/16 * 1/6 * 1/2)}
\label{table:table3}
\end{table}

Table \ref{table:table3} shows the performance of our model on the second setup of Social Relationship Graph Generation, \textit{SRGGen}. We achieve an accuracy of 27.64\% using our final model. The accuracy for the \textit{Our SRG-GN without scene} is 7.4\% lower than \textit{Our SRG-GN}, which empirically proves that context information plays a major role in generating a coherent social relationship graph.

\textbf{Results on PISC dataset:} Table \ref{table:piscmap} compares the mean-average precision evaluated on the PISC dataset for Social Relationship Recognition (SRRec). Our final model with mean pooling and 2 time steps notably outperforms the state-of-the-art model on PISC dataset by $\sim$8.5\%. Our final model improves only slightly in precision over our SRG-GN model without scene. One possible reason is that the scene context in PISC dataset has similar contextual information for the relationships unlike in the PIPA-relation graph dataset.

We report the precision of each of the 6 relationship labels in Table \ref{table:pisc_ap}. Our SRG-GN model improves in precision over the MN-CNN-only model for the classes \textit{couple} and \textit{commercial}. The class \textit{friends} has lower precision, indicating that other classes are sometimes wrongly classified as ``friends''. Due to imbalance in the training dataset, we introduce a weighted cross entropy loss to penalize the classes with few samples; this improves performance significantly.  

\begin{figure}[!tb]
\begin{center}
\includegraphics[width=0.9\linewidth]{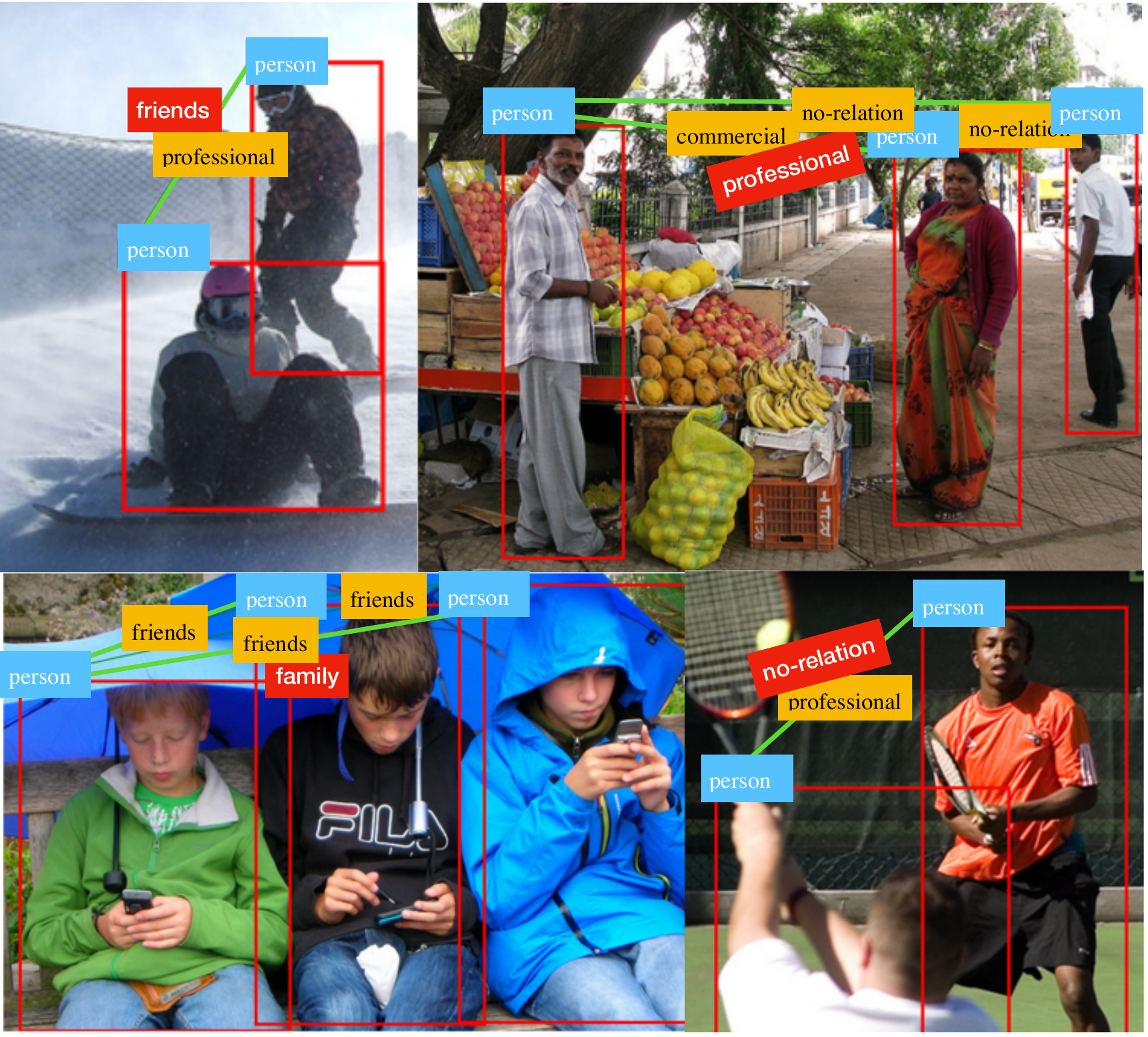}
\end{center}
\caption{Wrong relationship predictions from the SRG-GN model on the PISC dataset. The relationships in yellow are the ground-truth, the relationships in red are the incorrect predictions. Only the relationships marked as red in an image are incorrectly predicted by our model.}
\label{fig:pisc_fps}
\end{figure}

\begin{figure*}[!tb]
\begin{center}
\includegraphics[width=0.99\linewidth]{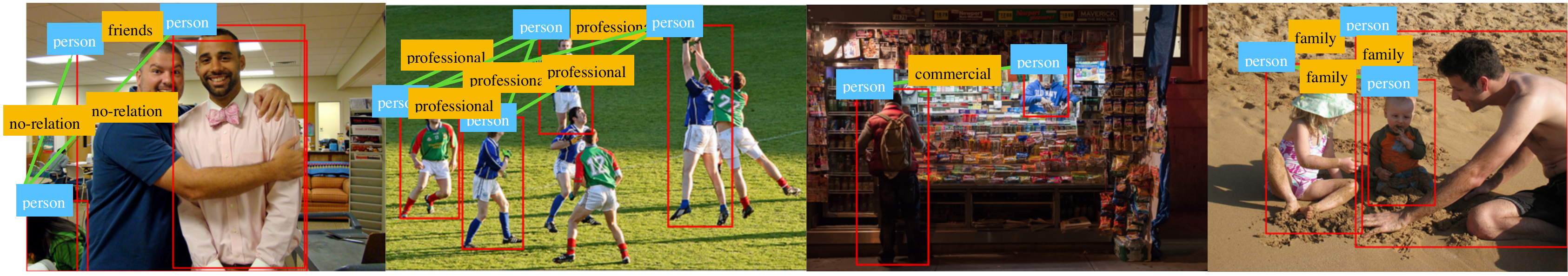}
\end{center}
\caption{Correct predictions from our final model on the PISC dataset.}
\label{fig:pisc_tps}
\end{figure*}

\subsubsection{Qualitative Results}
The Social Relationship graph (SRG) is a rich semantic graph with attribute and relationship information for the people in a given scene. Our SRG contains ground-truth information about the class and bounding-box labels of the objects in the image. Through our SRG-GN, we predict the social relationships, age and gender attributes of the people in a given scene. 

Figure \ref{fig:SRG_egs} shows qualitative results on PIPA-relation graph dataset to compare the SRG generated from our model and the ground truth. In the first example, the SRG-GN correctly predicts the relationships between the given people. As shown in the graph, all nodes (persons) have ``friends'' relationship between them which are correctly predicted by our model. Gender attributes also correspond to the ground-truth, but the age attributes are incorrectly predicted as ``middle-age''  instead of ``young-adult''. The model correctly predicts more complex relationships like ``sports-team members'' which has a lot more contextual information than other relationships like ``grandma-grandchild'' which it falsely predicts as ``mother-child'' due to ambiguity in such relationships.

Figure \ref{fig:pisc_tps} gives examples of the correct predictions on PISC dataset. Our model predicts multiple relationship instances in an image, such as a group of players are correctly labeled as ``professional''. Figure \ref{fig:pisc_fps} shows examples for misclassified relationships. For instance, the model falsely detects the relationship in bottom-left image as ``family'', when they are more likely to be friends due to information from adjacent nodes and edges. There is ambiguity between ``professional'' and ``commercial'' in some cases due to similar global and scene context for these classes.

\begin{table}[!tb]
\begin{center}
\centering
\resizebox{0.8\columnwidth}{!}{
    \begin{tabular}{|c|c|}
    \hline
    MODEL & mAP\\
    \hline\hline
    Pair--CNN+BBox & 54.3\%  \\
    Pair--CNN+BBox+Union & 56.9\% \\
    Pair--CNN+BBox+Global & 54.6\%  \\
    Pair--CNN+BBox+Scene & 51.7\% \\
    Dual-Glance & 63.2\% \\ \hline
    Our MN--CNN module only & 60.2\% \\
    Our SRG--GN without Scene & \textbf{69.2\%} \\
    Our SRG--GN (final model)  & \textbf{71.6\%} \\ 
    \hline
    \end{tabular}
}  
\end{center}
\caption{Mean--Average Precision (mAP) for the task of Social Relationship Recognition (\textit{SRRec}) on PISC dataset.}
\label{table:piscmap}
\end{table}

\begin{figure}[!tb]
\begin{center}
\includegraphics[width=\linewidth]{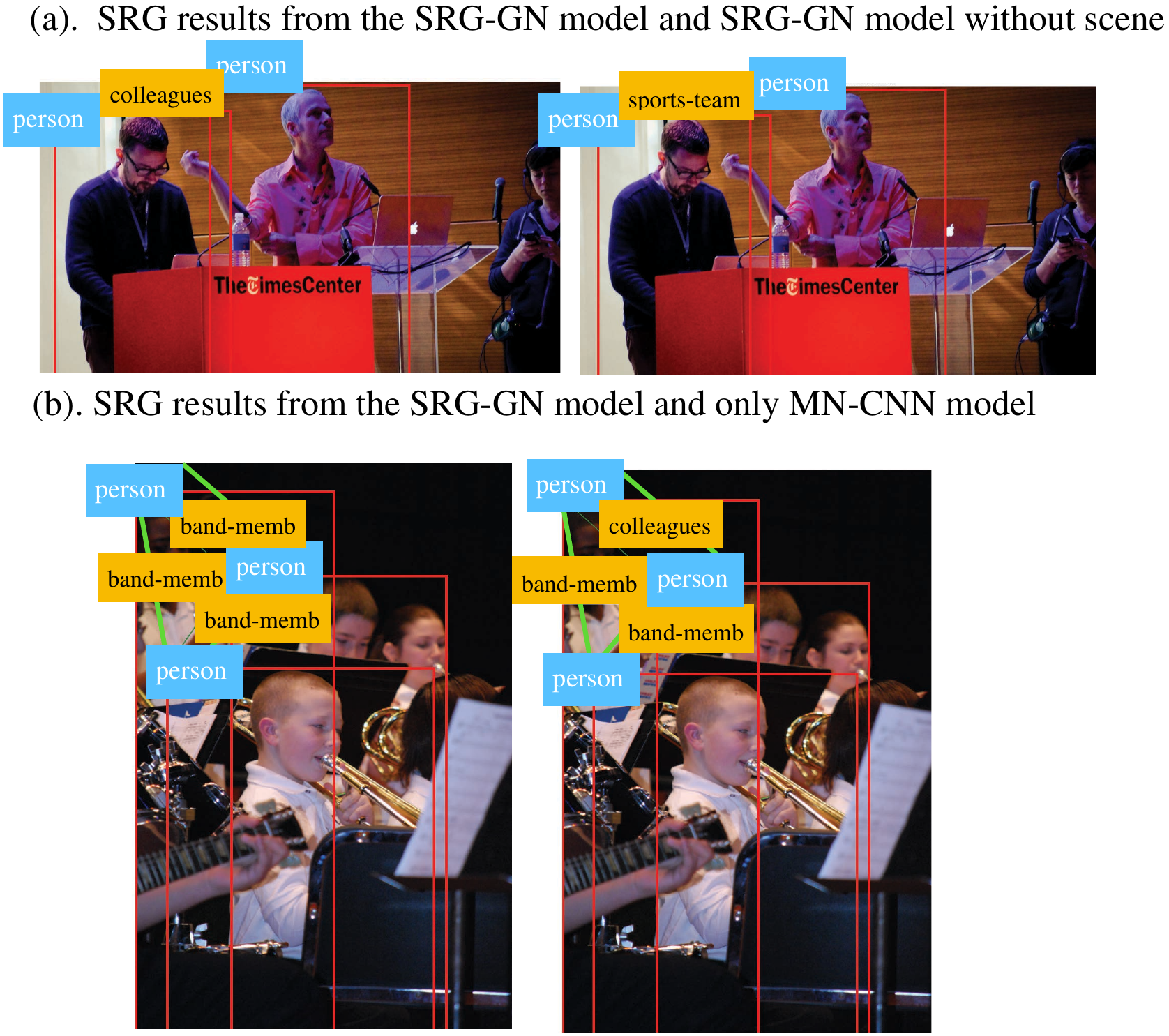}
\end{center}
\caption{Qualitative analysis of our model variations on PIPA-Relation. The left results are from our final model, \textit{SRG-GN}. The top-right result is from \textit{SRG-GN without Scene}, while the bottom-right result is from the \textit{only MN-CNN model}.}
\label{fig:Ablative}
\end{figure}

\section{Ablative Analysis}
In this section, we examine the performance of our \textit{SRG-GN} model variations on the PIPA-Relation graph dataset.
\subsection{Model Variations}

We evaluate the importance of scene context in predicting relationships in our final graph inference framework. As shown in Section \ref{results}, adding scene context significantly improves the performance on both tasks of \textit{SRRec} and \textit{SRGGen}. Intuitively, we can infer that scene information can be important in many different situations. For instance, given a party scene, the group of people are more likely to be friends than colleagues, and a group of athletes running on a track are much more likely to be sports team members than band members. In Figure \ref{fig:Ablative}(a), we present an example to highlight the importance of using whole image scene context for accurate predictions. Our \textit{SRG-GN without scene} incorrectly predicts the two people as sports team members, but if we look at the whole scene together it increases the chances of them being colleagues and not related to sports. Without scene context, identifying the relationships between two people can be sometimes ambiguous. This clearly explains the motivation behind using scene context as an important feature in the \textit{SRG-IN} module.

We also examine how predicting relationships in isolation from the \textit{only MN-CNN} module has lower accuracy than the combined model with the SRG-IN module. For example, a group of people performing on the stage should all very likely be band members, and our model exploits this information for overall inference, whereas the \textit{only MN-CNN} module predicts the triplets in the social relationship graph independently. In Figure \ref{fig:Ablative}(b), our final model correctly predicts the relationships as band-members due to the message information from the adjacent group of relationships in an image. Without this message passing network, the MN-CNN module only considers information from the pair of people between whom relationship has to be predicted. Thus, the \textit{SRG-IN} module uses contextual information from the nearby nodes and edges in a graph to improve individual predictions.
\begin{table}[!tb]
\begin{center}
\centering
\resizebox{0.7\columnwidth}{!}{
    \begin{tabular}{|c|c|c|}
    \hline
    Pooling & \# time steps & Accuracy \\
    \hline\hline
    max & 1 & 50.41\%  \\
    \textbf{max} & \textbf{2}  & \textbf{52.16\%} \\
    max & 3 & 51.27\%  \\ \hline
    mean & 1 & 50.89\% \\ 
    \textbf{mean} & \textbf{2} & \textbf{53.56\%} \\
    mean & 3 & 52.08\% \\
    \hline
    \end{tabular}
    }
\end{center}
\caption{Ablation study for different time--steps and pooling techniques on the PIPA-relation graph dataset.}
\label{table:pool}
\end{table}

\subsection{Pooling and Time--Step variations}
\label{pooling}
We evaluate our SRG-GN model on the PIPA-relation with different number of time steps and pooling techniques. From Table \ref{table:pool}, it can be observed that mean-pooling is more effective in passing useful information between hidden states than max-pooling. Also, there is a $\sim$1.5\% decrease in accuracy on increasing the time steps as it starts passing noisy information between states with more false detections in the social relationship graph.

\section{Conclusion}
We introduced a novel end-to-end-trainable network for generating social relationship graphs from images using GRUs. Previous work on generating graphs dealt with relationships between objects, whereas our work tackles the more challenging problem of inferring social relationships. Experimental results show the importance of using attribute and contextual features with message passing in a graph. Our model outperforms the state-of-the-art for recognizing social relationships, and performs well for generating social relationship graphs. This work can be extended for more complex tasks, such as predicting social intentions.

{\small
\section*{Acknowledgements}
This work was supported by NRF grant no. NRF2015-NRF-ISF001-2541 (KTM and CT) and A*STAR SERC SSF grant no. A1718g0048 (AG and KTM).
}
{\small
\bibliographystyle{ieee}
\bibliography{egbib}
}

\end{document}